\documentclass[runningheads]{llncs}
\usepackage{graphicx}
\usepackage{amsmath}
\usepackage{amssymb}
\usepackage{bm}
\usepackage{subfigure}
\usepackage{booktabs}
\usepackage{multirow}
\usepackage{booktabs}
\usepackage{caption}
\usepackage{float}
\usepackage{algorithmic}
\usepackage{algorithm}
\usepackage{bbding}
\usepackage{array}
\usepackage{color}
\usepackage{diagbox}
\usepackage{threeparttable}
\usepackage{pifont}
\usepackage{tikz}
\usepackage{amsmath,amssymb} 
\usepackage{color}
\usepackage{amssymb}%
\usepackage{comment}
\usepackage[accsupp]{axessibility}  %
\usepackage{bbding}

\def\etal{\emph{et al.}}
\def\ie{\emph{i.e.}}
\begin{document}

\title{Disentangling 3D Attributes from a Single 2D Image: Human Pose, Shape and Garment} 

\titlerunning{Disentangling 3D Attributes from a Single 2D Image}

\author{Xue Hu\textsuperscript{2$\star$} \hspace{4mm}
    Xinghui Li\textsuperscript{3$\star$} \hspace{4mm}
    Benjamin Busam \textsuperscript{4}\hspace{4mm}
    Yiren Zhou\textsuperscript{1} \\
    Ales Leonardis\textsuperscript{1}\hspace{4mm}
    Shanxin Yuan\textsuperscript{1$\dagger$}}
\authorrunning{X. Hu, X. Li, B. Busam, Y. Zhou, A. Leonardis, S. Yuan}
\institute{\footnotesize{$^1$Huawei Noah's Ark Lab \hspace{5mm} $^2$Imperial College London}  \\ \footnotesize{$^3$University of Oxford \hspace{5mm}  $^4$Technical University of Munich} }

\maketitle
\def\thefootnote{*}\footnotetext{Equal contribution}
\def\thefootnote{$\dagger$}\footnotetext{Corresponding author}

\begin{abstract}
For visual manipulation tasks, we aim to represent image content with semantically meaningful features. However, learning implicit representations from images often lacks interpretability, especially when attributes are intertwined. We focus on the challenging task of extracting disentangled 3D attributes only from 2D image data. Specifically, we focus on human appearance and learn implicit pose, shape and garment representations of dressed humans from RGB images. Our method learns an embedding with disentangled latent representations of these three image properties and enables meaningful re-assembling of features and property control through a 2D-to-3D encoder-decoder structure. The 3D model is inferred solely from the feature map in the learned embedding space. To the best of our knowledge, our method is the first to achieve cross-domain disentanglement for this highly under-constrained problem. We qualitatively and quantitatively demonstrate our framework's ability to transfer pose, shape, and garments in 3D reconstruction on virtual data and show how an implicit shape loss can benefit the model's ability to recover fine-grained reconstruction details.
\end{abstract}

\section{Introduction}

If you reconstruct a 3D model from a single image, you also want the power to control its content in a meaningful way. This wish has created a wide range of applications that leverage learning of implicit representations by disentanglement,
such as face identity swapping \cite{gao2021information,zhang2019multi}, hairstyle transfer \cite{sanchez2020learning,higgins2016beta} and pose and shape transfer \cite{zhou20unsupervised,Jiang2020HumanBody,mu2021sdf}. The main objective of implicit representation learning is to disentangle and encode information regarding each characteristic of input signals such that a new sample can be generated by manipulating learned representations. For example, if we separate pose and shape information from the 3D model of a person, we can achieve pose transfer by simply replacing the pose information while keeping the original shape information.

However, the current literature only discusses cases where input and output are from the same domain. This means that if the input is an image or a 3D mesh, the output would also take the same form as the input and learned representations can only control the output in the same domain. Due to this restriction, current works are difficult to be directly applied to 2D-to-3D tasks such as Augmented Reality (AR), where 3D information is often inferred from easily acquired 2D images~\cite{jung2022my,wang2022phocal,gao2022ppp,busam2020like,manhardt2020cps++}. 

In this paper, we propose a solution to this highly under-constrained problem. We focus on learning pose, shape and garment representations of dressed human bodies from 2D RGB images and use these representations to manipulate corresponding 3D models. Our inspiration is drawn from the fact that the 3D mesh of a dressed human can be solely estimated from the feature map of its images \cite{saito2019pifu,saito2020pifuhd} by training a shape prior. A key deduction from this observation is that the feature map of an object's 2D signal contains sufficient information to construct its 3D model if a shape prior is provided. Therefore, if we disentangle and reconstruct feature maps of input signals rather than input signals themselves, we can subsequently control the final 3D models which are inferred from the feature maps using the shape prior. 

\begin{figure*}[t]
    \centering
    \includegraphics[width=0.95\textwidth]{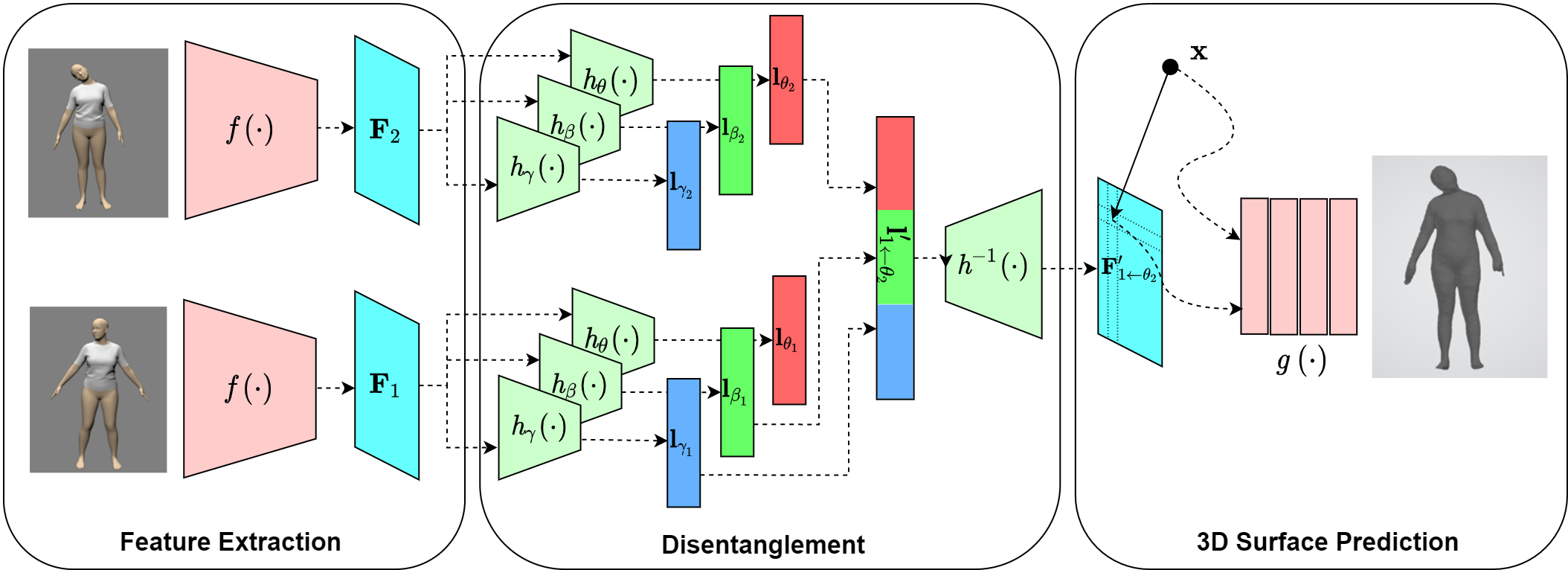}
    \caption{This figure illustrates the architecture of our method. The feature map is $\mathbf{F}$ is extracted from the image by the feature extractor $\mathit{f}(\cdot)$. The feature map is then disentangled into three latent codes $\mathbf{l}_{\theta}$, $\mathbf{l}_{\beta}$ and $\mathbf{l}_{\gamma}$ which correspond to pose, shape and garment representations, respectively.
    By swapping the latent code, a feature map $\mathbf{F}^{\prime}$ with swapped information is recovered by $\mathit{h}^{-1}(\cdot)$, from which the 3D mesh with swapped property is inferred as the implicit expression by $\mathit{g}(\cdot)$, where $\mathit{g}(\cdot)$ takes both a 3D coordinate $\mathbf{x}$ and the aligned feature of the projection of $\mathbf{x}$ on the image. 
    }
    \label{fig:pipeline}
\end{figure*}

Our method consists of three parts: a feature extractor, a multi-head encoder-decoder and an MLP, as illustrated in Fig. \ref{fig:pipeline}.
A feature extractor firstly extracts a feature map from the input image. Further, a consecutive feature extractor learns an embedding that encodes disentangled representations for pose, shape and garment into three respective latent codes. The feature map is then recovered from the latent codes, and finally, an MLP is used as a shape prior to construct the 3D model based on the pixel-aligned feature interpolated from the feature map. To change the pose, shape or garment of the 3D model individually while keeping the other properties, we can change the corresponding latent code, which consequently changes the generated 3D mesh. 

To the best of our knowledge, our model is the first method to use representations learned from 2D input to control 3D output. Although we exemplify the power of the method with dressed human bodies, such a principle can be generalised to any class of object if the shape prior can be properly trained. In contrast to standard human modelling methods \cite{joo2018total,loper2015smpl,pons2015dyna,anguelov2005scape}, we are able to control the model without using a template. To summarize, our contributions are twofold:
\begin{itemize}
    \item We propose a method to disentangle 3D shape attributes from 2D image data. We exemplify the principle by learning pose, shape and garment representations of dressed humans from 2D images and allow expressive control of reconstructed 3D model from disentangled feature sub-manifolds. Our method is the first to achieve 2D-to-3D representation learning and output manipulation. 
    
    \item We analyse design choices and provide experimental evidence of controlled feature manipulation for pose, shape and garment representations from 2D input on a publicly available dataset \cite{patel20tailornet}. 
\end{itemize}

\section{Related Works}
\label{sec:relate}

\paragraph{Disentanglement} The objective of disentanglement is to find underlying latent representations that control the variation of the data. The pioneering work in this area is InfoGAN \cite{chen2016infogan} which is based on Generative Adversarial Networks (GANs) \cite{goodfellow2014generative} and aims to maximize the mutual information between the latent codes and generator distribution. $\beta$-VAE~\cite{higgins2016beta} and its variant \cite{chen2018isolating} use a Variational Autoencoder (VAE) \cite{kingma2013auto} rather than GAN model and penalize a KL-Divergence term to enhance the independence within latent space.

Disentanglement can be applied to both 2D image inputs and 3D inputs, to achieve either attribute transfer like face identity swapping \cite{gao2021information,zhang2019multi} and hairstyle transfer \cite{sanchez2020learning,higgins2016beta} on 2D images, or pose and shape transfer \cite{zhou20unsupervised,Jiang2020HumanBody,mu2021sdf} on 3D inputs. Our method is more related to \cite{zhou20unsupervised} where pose and shape representations are learned in an unsupervised manner. The most significant differences between our method and most existing pipelines are that our input and output are from different domains (2D input and 3D output), and we disentangle the image's feature map which is an abstract representation rather than the raw data. These differences make our task much more challenging. 

\paragraph{Parametric Human Modelling}  Parametric models targeting humans initially focus on the parameterization of the naked human body. Various body templates have been proposed in the past decade \cite{osman2020star,loper2015smpl,anguelov2005scape,hasler2009statistical}. Such a parameterization has recently been generalized to the modelling of the garment \cite{bhatnagar2019mgn,patel20tailornet,santesteban2021self,ma2020cape,ma2021scale,gundogdu2019garnet,lahner2018deepwrinkles}. Compared to the naked human body, the parameterization of the garment is much more difficult due to the complicated local details such as wrinkles and foldings. The garment models are usually associated with the naked body model such as SMPL. They are either represented as separated SMPL-like templates \cite{patel20tailornet,santesteban2021self,zhu2020deep} or displacement fields to the body model \cite{ma2020cape,alldieck2019learning}. Some works \cite{patel20tailornet,bhatnagar2019mgn} additionally model the intra-class garment variation using statistical tools such as principle component analysis (PCA).

\paragraph{Implicit Neural Representation} Implicit neural representation aims to represent a 3D object using an implicit function which typically takes the form of an MLP. The pioneering works are OccNet \cite{Occupancy_Networks} and DeepSDF \cite{Park_2019_CVPR} which encode the objects as an occupancy field and a signed distance field, respectively. Compared with traditional 3D representation such as voxel or mesh, neural representation allows a continuous surface representation which circumvents the loss of accuracy due to discretization. Following these two works, many variations of implicit functions \cite{sitzmann2020implicit,Peng2020ECCV,jiang2020local,chabra2020deep} and training losses \cite{gropp2020igr} have been proposed. The aforementioned methods mainly focus on 3D input such as point clouds, but PiFU \cite{saito2019pifu} generalizes the implicit function to 2D images by proposing a pixel-aligned implicit function. Instead of a 3D coordinate, the function takes in the corresponding 2D image feature of the 3D location and the depth value, and it can infer a wide variety of human poses and shapes. DISN \cite{xu2019disn} also has a similar structure to construct static objects. Our method is closely related to PiFU as we disentangle pose and shape information from the image's feature map. 

\section{Method}
\label{sec:method}
Our method intends to disentangle pose, shape and garment representations of a dressed human from a single RGB image and simultaneously infer the 3D mesh of it. Existing works \cite{saito2019pifu,saito2020pifuhd} have already demonstrated that 3D mesh can be inferred solely from a single image by pixel-aligned implicit function based on the feature map of the image. This indicates that such a feature map contains all information regarding the target mesh; hence, pose, shape and garment representation can be disentangled. Therefore, our method consists of three parts: a feature extractor that extracts the feature map from the image, a multi-head encoder-decoder that disentangles and reconstructs the feature map, and an MLP which infers the 3D mesh from the feature map. We illustrate this pipeline in Fig. \ref{fig:pipeline}.

\subsection{Feature Disentanglement}
\label{disentangle}
After feature extraction, the feature disentanglement is achieved by a multi-head autoencoder module. Given an image $\mathbf{I}_{1}$, its feature map $\mathbf{F}_{1} \in \mathbb{R}^{H\times W\times C}$ can be extracted using the feature extractor $\mathit{f}(\cdot)$. Such a feature map encodes all the necessary information to construct a 3D model, as the missing depth information would be compensated by the implicit function, which acts like a 3D human shape prior \cite{saito2019pifu}. Hence, we only need to disentangle the feature map so that a new feature map and thus the 3D human body can be reconstructed after latent editing using the encoded information from an image of the desired human model.

The disentanglement network follows a similar design to the work of Zhou \etal~\cite{zhou20unsupervised}. To disentangle the pose, shape and garment properties, the feature map $\mathbf{F}_{1}$ is past through three separated encoder modules $\mathit{h}_{\theta}(\cdot)$, $\mathit{h}_{\beta}(\cdot)$ and $\mathit{h}_{\gamma}(\cdot)$, which downsample, flatten and project the feature map into three latent codes: $\mathbf{l}_{\theta_{1}}$, $\mathbf{l}_{\beta_{1}}$ and $\mathbf{l}_{\gamma_{1}}$, which represent pose, shape and garment information of the 3D model, respectively. With three latent codes, the feature map can be reconstructed by concatenating the latent codes together into a longer latent vector $\mathbf{l}^{\prime}_{1}$, where $\mathbf{l}^{\prime}_{1} = \texttt{cat}[\mathbf{l}_{\theta_{1}}$, $\mathbf{l}_{\beta_{1}}$, $\mathbf{l}_{\gamma_{1}}]$. Such a vector is then passed through the decoder module $\mathit{h}^{-1}(\cdot)$ of the encoder-decoder, which upsamples the vector and then reconstructs the feature map $\mathbf{F}^{\prime}_{1}$. The pose, shape or garment transfer can be achieved by simply replacing the corresponding latent code. For example, if we want to transfer the pose of the person in $\mathbf{I}_{1}$ to another pose of the person in $\mathbf{I}_{2}$, we just need to extract the latent codes $\mathbf{l}_{\theta_{2}}$, $\mathbf{l}_{\beta_{2}}$ and $\mathbf{l}_{\gamma_{2}}$ from $\mathbf{I}_{2}$ and replace the latent code $\mathbf{l}_{\theta_{1}}$ with $\mathbf{l}_{\theta_{2}}$. As a result, the feature map $\mathbf{F}^{\prime}_{1\leftarrow \theta_{2}}$ reconstructed from concatenated latent code $\mathbf{l}^{\prime}_{1\leftarrow \theta_{2}} = \texttt{cat}[\mathbf{l}_{\theta_{2}}$, $\mathbf{l}_{\beta_{1}}$, $\mathbf{l}_{\gamma_{1}}]$ would contain the shape and garment information of $\mathbf{I}_{1}$ but the pose information of $\mathbf{I}_{2}$. According to $\mathbf{F}^{\prime}_{1\leftarrow \theta_{2}}$, the implicit surface expression can be predicted by the surface prediction MLP.

\subsection{3D Reconstruction}
\label{3dRecon}
In recent years, the implicit neural representation has exhibited the exceptional capability in 3D rendering \cite{saito2021scanimate,sitzmann2020implicit,saito2019pifu,Park_2019_CVPR}. The object or scene is often encoded into an MLP as an occupancy field (OCC) or a signed distance field (SDF). Mathematically, it is defined as:
\begin{equation}
    \label{occ_eq}
    s_{\mathbf{x}} = \mathit{g}(\mathbf{x})
\end{equation}
where $\mathit{g}(\cdot)$ is the MLP and $s_{\mathbf{x}}$ is the learned implicit surface expression at the sampling position $\mathbf{x}\in \mathbb{R}^{3}$. The object's surface is represented by the level set of the implicit function $\mathit{g}(\cdot)$, from which the 3D geometry can be easily constructed by Marching Cube. Normally, a MLP could only store one scene or object \cite{saito2021scanimate,sitzmann2020implicit,Park_2019_CVPR}. However, a single MLP $\mathit{g}(\cdot)$ can be generalized to multiple objects by taking feature map of the image as an input \cite{saito2019pifu,saito2020pifuhd}:
\begin{equation}
    \label{pifu_eq}
    s_{\mathbf{x}} = \mathit{g}(\mathbf{F}_{x}, \mathbf{x}_{z})
\end{equation}
where $x$ is the projection of $\mathbf{x}$ on to the image plane, $\mathbf{F}_{x}\in \mathbb{R}^{C}$ is the value of encoded feature map $\mathbf{F}$ at a sampling point $x$, and $\mathbf{x}_{z}$ is the depth value of $\mathbf{x}$. By taking the pixel-aligned feature $\mathbf{F}_{x}$, multiple objects can be inferred from a single MLP and objects can be controlled by editing the feature map.

For 2D-3D human reconstruction task, the formulation in Eq. \ref{pifu_eq} is commonly supervised using the ground truth occupancy values \cite{saito2019pifu}. However, compared with occupancy, we consider the SDF as a better supervision signal during training because using SDF we can not only supervise the distance value but also the surface orientation. Unfortunately, the formulation in Eq. \ref{pifu_eq} is not suitable for SDF supervision as the output is not differentiable to the $\mathbf{x}$ hence the surface norm is unable to be computed. In order to supervise the surface norm, and being inspired by \cite{yu2021pixelnerf}, we change the formulation in Eq. \ref{pifu_eq} into:
\begin{equation}
    \label{sdf_eq}
    s_{\mathbf{x}} = \mathit{g}(\mathbf{F}_{x}, \sigma(\mathbf{x})) ~ \hat{=}~ \mathit{g}(\mathbf{F}, \mathbf{x})
\end{equation}
where $\sigma(\cdot)$ is the positional encoding function defined in \cite{yu2021pixelnerf}.
For the sake of simplification, we denote the MLP expression $\mathit{g}(\mathbf{F}_{x}, \sigma(\mathbf{x}))$ as $\mathit{g}(\mathbf{F}, \mathbf{x})$. At here, we explicitly take the 3D location $\mathbf{x}$ as an input and hence the surface norm can be computed and supervised. 

\subsection{Loss}
\label{loss}
Our loss includes two parts: reconstruction loss and disentanglement loss.

\paragraph{3D Reconstruction Loss} In order to learn the feature disentanglement from 2D images, we need to first initialize the feature extractor $\mathit{f}(\cdot)$ and the MLP modules $\mathit{g}(\cdot)$ for a 3D shape prior. $\mathit{f}(\cdot)$ followed by $\mathit{g}(\cdot)$ makes up the direct reconstruction pipeline similar to \cite{saito2019pifu}, which can be supervised by the 3D reconstruction loss according to ground-truth 3D meshes $\mathbf{M}$. We define the reconstruction loss as $\mathcal{L}_{recon}(\mathbf{F}, \mathbf{X}, \mathbf{M})$, where $\mathbf{X}$ is a set of 3D points that are sampled around the surface of $\mathbf{M}$ (details could be found in supplementary materials). We explore two implicit representations: occupancy and signed distance function (SDF). For occupancy, we use the same reconstruction loss as defined in \cite{saito2019pifu}. For SDF,  we use the same reconstruction loss as defined in \cite{saito2021scanimate,sitzmann2020implicit}. More details are provided in the supplementary material. 

\paragraph{Disentanglement Loss} With the properly pre-trained feature extractor and MLP surface predictor, we then jointly train the encoder-decoder for feature disentanglement and MLP for surface refinement. 
In order to properly learn the disentanglement of pose, shape and garment, we adopt the control variate strategy. We construct a dataset which consists of pairs of images and their corresponding ground-truth 3D meshes. For each pair, the difference between the images is either in pose, shape or garment, while the other two are kept same. This control strategy allows the encoder-decoder to focus on one property at each time and hence interpret the difference in latent space. We provides more details about our dataset pairing in section \ref{dataset}. 

For each image pair, denote the different property between them as $d$, where $d\in \{\theta, \beta, \gamma\}$ and two remaining properties as $s1$ and $s2$, where $s1,s2 \in \{\theta, \beta, \gamma\}\setminus \{d\}$. Let $\mathbf{F}_{1} = \mathit{f}(\mathbf{I}_{1})$ and $\mathbf{F}_{2} = \mathit{f}(\mathbf{I}_{2})$, then $\{\mathbf{l}_{s1,1}, \mathbf{l}_{s2,1}, \mathbf{l}_{d1}\} = h(\mathbf{F}_{1})$ and $\{\mathbf{l}_{s1,2},$ $\mathbf{l}_{s2,2}, \mathbf{l}_{d2}\} = h(\mathbf{F}_{2})$. We define two procedures:
\begin{itemize}
\item \textit{Disentangled self-reconstruction} If no latent code swapping is applied, we should be able to reconstruct the feature maps $\mathbf{F}_{1}$ and $\mathbf{F}_{2}$ by $\mathbf{F}^{\prime}_{1} = h^{-1}(\mathbf{l}^{\prime}_{1})$ and $\mathbf{F}^{\prime}_{2} = h^{-1}(\mathbf{l}^{\prime}_{2})$, where $\mathbf{l}^{\prime}_{1} = \texttt{concat}[\mathbf{l}_{s1,1},\mathbf{l}_{s2,1}, \mathbf{l}_{d1}]$ and  $\mathbf{l}^{\prime}_{2} = \texttt{concat}[\mathbf{l}_{s1,2},$ $ \mathbf{l}_{s2,2}, \mathbf{l}_{d2}]$.
We should also be able to recover $\mathbf{M}_{1}$ and $\mathbf{M}_{2}$ from the reconstructed feature map $\mathbf{F}^{\prime}_{1}$ and $\mathbf{F}^{\prime}_{2}$.  
\item \textit{Disentangled cross-reconstruction} In the case where the latent codes of property $d$ is swapped, the modified latent codes become: $\mathbf{l}^{\prime}_{1\leftarrow d2} = \texttt{concat}[\mathbf{l}_{s1,1}, $ $\mathbf{l}_{s2,1}, \mathbf{l}_{d2}]$, then two new feature maps can be reconstructed as $\mathbf{F}^{\prime}_{1\leftarrow d2}=h^{-1}(\mathbf{l}^{\prime}_{1\leftarrow d2})$ and $\mathbf{F}^{\prime}_{2\leftarrow d1}=h^{-1}(\mathbf{l}^{\prime}_{2\leftarrow d1})$. Since there is only one different property between two images, the reconstructed feature map of one image should be the same as the original feature maps of the other and so are their 3D meshes. 
\end{itemize}

We define the feature map reconstruction loss before and after disentanglement during self- and cross-reconstruction:
\begin{multline}
    \mathcal{L}_{feat} = \| \mathbf{F}_{1}-\mathbf{F}^{\prime}_{1} \|_2^2 + \| \mathbf{F}_{2}-\mathbf{F}^{\prime}_{2} \|_2^2
    + \| \mathbf{F}_{1}-\mathbf{F}^{\prime}_{2\leftarrow d1} \|_2^2 + \| \mathbf{F}_{2}-\mathbf{F}^{\prime}_{1\leftarrow d2} \|_2^2 
\end{multline}

We also add the latent identity loss $\mathbf{L}_{latent}$ which matches the compressed latent code of two invariant properties in each image pair:
\begin{equation}
    \mathbf{L}_{latent} = \|\mathbf{l}_{s1, 1}-\mathbf{l}_{s1, 2}\|_2^2 + \|\mathbf{l}_{s2, 1}-\mathbf{l}_{s2, 2}\|_2^2
\end{equation}

We define a surface reconstruction loss (either with occupancy or SDF) between the ground truth mesh $\mathbf{M}$ and the predicted output $\mathit{g}(\cdot)$. This term direct supervises the quality of the 3D models generated from the reconstructed feature maps:
\begin{multline}
    \mathcal{L}_{recon} = \mathcal{L}_{recon}(\mathbf{F}^{\prime}_{1}, \mathbf{X}_{1}, \mathbf{M}_{1}) + \mathcal{L}_{recon}(\mathbf{F}^{\prime}_{2}, \mathbf{X}_{2}, \mathbf{M}_{2}) \\
    + \mathcal{L}_{recon}(\mathbf{F}^{\prime}_{2\leftarrow d1}, \mathbf{X}_{1}, \mathbf{M}_{1}) + \mathcal{L}_{recon}(\mathbf{F}^{\prime}_{1\leftarrow d2}, \mathbf{X}_{2}, \mathbf{M}_{2})
\end{multline}

Therefore, for each image pair, the overall disentanglement loss $\mathcal{L}_{disent}$ is defined as the weighted combination of the above three terms:
\begin{equation}
    \mathbf{}\mathcal{L}_{disent} = \mathcal{L}_{feat}+\mathcal{L}_{latent}+\mathcal{L}_{recon}
\end{equation}

\section{Experiment}
\label{sec:exp}

\subsection{Dataset}
\label{dataset}

It is extremely challenging to obtain a real-world dataset that contains two humans with different shapes and garments but in the same pose. Therefore, we modified a public physically simulated dataset, TailorNet dataset \cite{patel20tailornet} to evaluate our method. The original dataset consists of physically simulated 3D sequences of motions of dressed humans, which are modelled by the SMPL model with an additional garment layer on top of it. The dataset includes humans with a range of body shapes and several types of garments, where each type of garment also has intra-class variation. We make the following modifications to the dataset: (i) We construct the image pairing by comparing the SMPL parameters of the human body and garment type. In the end, we obtain 1369 pose-vary pairs, 51 shape-vary pairs and 761 garment-varying pairs. (ii) Since the combined ground-truth meshes of dressed humans have multiple layers, the normals are in the opposite direction at overlapping regions. This may confuse the network during training with SDF-based supervision loss. Therefore, we convert them into single-layer water-tight meshes using the Manifold package. (iii) For each mesh, we render the front view using Blender as the input to our method. The final output is the three subsets that contain pairwise data variation. The three subsets are padded to the same size of 1369 and split into the training, validation and test dataset with a ratio of 6:2:2.

\begin{table}
\begin{center}
\begin{tabular}{|l|llll|}
\hline
Experiment       &   Implicit Func            & Chamfer (mm) & P2S (mm) & normal (mm) \\ \hline \hline
self  & SDF/OCC & \textbf{1.43}/3.65  & \textbf{11.21}/22.24     &  11.16/\textbf{8.93}  \\ 
cross - pose & SDF/OCC & \textbf{1.26}/4.73 & \textbf{12.12}/21.63 &   10.63/\textbf{8.72} \\ 
cross - shape & SDF/OCC &  \textbf{2.46}/3.09 & \textbf{12.25}/19.88   &  11.46/\textbf{9.28} \\ 
cross - garment & SDF/OCC & \textbf{1.26}/3.36 &  \textbf{9.35}/22.61 &  11.53/\textbf{8.86} \\ \hline
\end{tabular}
\end{center}
\caption{Reconstruction error (median) after swapping the latent code of body shape, pose and garment style, using SDF and occupancy.}
\label{tab:swap}
\end{table}

\subsection{Implementation Details}
Our method is implemented using PyTorch. For the feature extractor $\mathit{f}(\cdot)$, we follow the choice in \cite{saito2019pifu} and use the Hourglass Network \cite{newell2016stacked}. For the encoder $\mathit{h}(\cdot)$ and decoder modules $\mathit{h}^{-1}(\cdot)$, we use a ResNet18 \cite{he2016deep} implemented in Bolts library~\cite{falcon2020framework}. We change the input dimension of the first convolutional layer to match the dimension of the feature map. The latent size for $\mathbf{l}_{\theta}$, $\mathbf{l}_{\beta}$ and $\mathbf{l}_{\gamma}$ are are all set to be 128 according to the ablation study (see details in supplementary material). For the surface predictor, we follow the design in \cite{yu2021pixelnerf} which is a fully connected MLP with residual link between every two linear layers. 

We first pre-train the feature extractor and surface predictor on the direct reconstruction pipeline. For both occupancy and SDF supervision, we use the RMSprop optimizer with a learning rate of $1\times10^{-4}$, the number of sampling is 1000 and batch size is 8. After epoch 150, the learning rate is reduced to $1\times10^{-5}$ to ensure smooth convergence. During the training for feature disentanglement, the feature extractor is frozen while the encoder-decoder and surface predictor are tuned together. The same optimizer is used except that the learning rate is kept as $1\times10^{-4}$ until the convergence reaches around 360 epochs. 

\subsection{Results}

\subsubsection{Reconstruction with latent representation}

\begin{figure*}
  \centering
\includegraphics[width=0.99\textwidth]{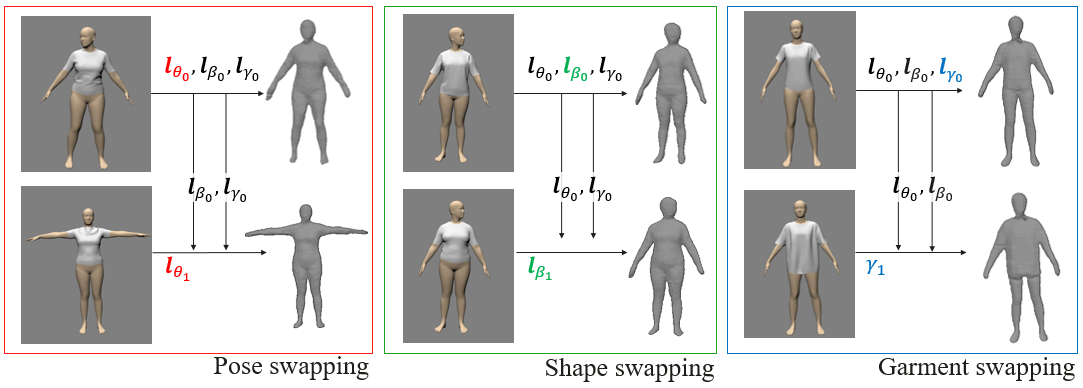}
    \caption{Example results of surface reconstruction after pairwise swapping the encoded latent components. Note that the pose and shape swapping are achieved on SDF-based pipeline, while the garment swapping is achieved on occupancy-based pipeline.}
    \label{fig:swap} 
\end{figure*}

Our proposed encoder-decoder structure allows us to disentangle and modify the latent code $\mathbf{l}$ by replacing its components $\mathbf{l}_{\theta}$, $\mathbf{l}_{\beta}$ and $\mathbf{l}_{\gamma}$. We test the quality of self-reconstruction and pairwise cross-reconstruction defined in Section \ref{loss}. The reconstruction quality is measured using three metrics: the Chamfer distance between reconstructed and ground-truth surfaces, the Euclidean distance from predicted vertices to their closest point on the ground-truth surface (P2S), and the RMS difference between the predicted and ground-truth surface normal vectors. We summarise the results in Table \ref{tab:swap}, where ``self '' means self-recon and ``cross'' means cross-recon.
The SDF-based pipeline achieves a higher accuracy than the occupancy-based pipeline. However, from visual inspections, we find the SDF-based method can better distinguish between different body poses and shapes than garment styles. By contrast, occupancy-based disentangled reconstruction capture all three variations, though the reconstruction quality is lower. The difference may be because the SDF approach tends to over-smoothen the predicted surface and the garment variation in the training dataset is too subtle. Fig. \ref{fig:swap} shows the visual demonstration for how the mesh reconstruction is controlled through pairwise latent swapping.

\begin{figure*}
  \centering
\includegraphics[width=\textwidth]{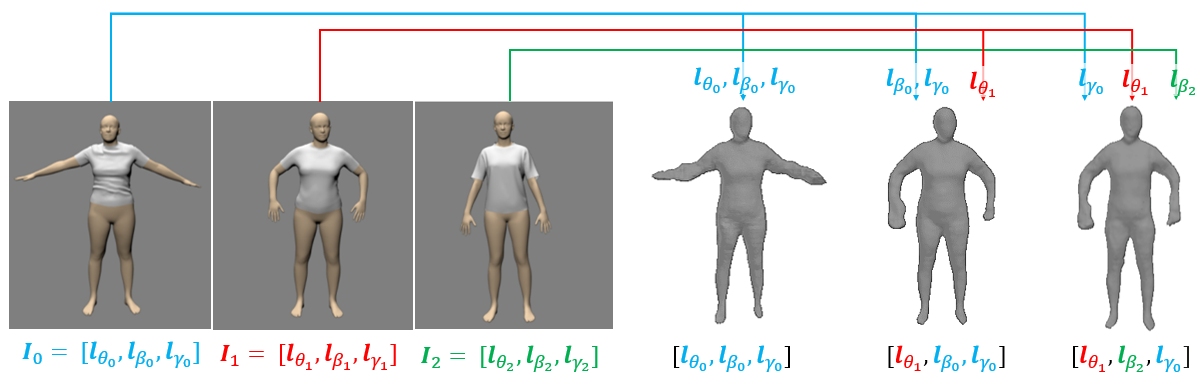}
    \caption{Example results of surface reconstruction after uncontrolled swapping of both body pose and shape. The synthetic mesh (rightmost) incorporates the tight garment style of $\mathbf{I}_{0}$, body pose of $\mathbf{I}_{1}$ and slimmer body shape of $\mathbf{I}_{2}$.}
    \label{fig:rand_swap} 
\end{figure*}

Pairwise swapping can only change one property each time and requires the other two to be identical, which is hardly possible in reality. Therefore, given a specific garment, we test the uncontrolled swapping where the body pose and shape latent code are replaced by new values at the same time. As shown in Fig. \ref{fig:rand_swap}, the body pose $\mathbf{l}_{\theta}$, shape $\mathbf{l}_{\beta}$ and garment information $\mathbf{l}_{\gamma}$ are respectively encoded from $\mathbf{I}_{1}$, $\mathbf{I}_{2}$ and $\mathbf{I}_{0}$, concatenated, and decoded into a new human body. We have included more examples in the supplementary material.

\subsubsection{Interpolation test}
\begin{figure}
  \centering
  \includegraphics[width=\textwidth]{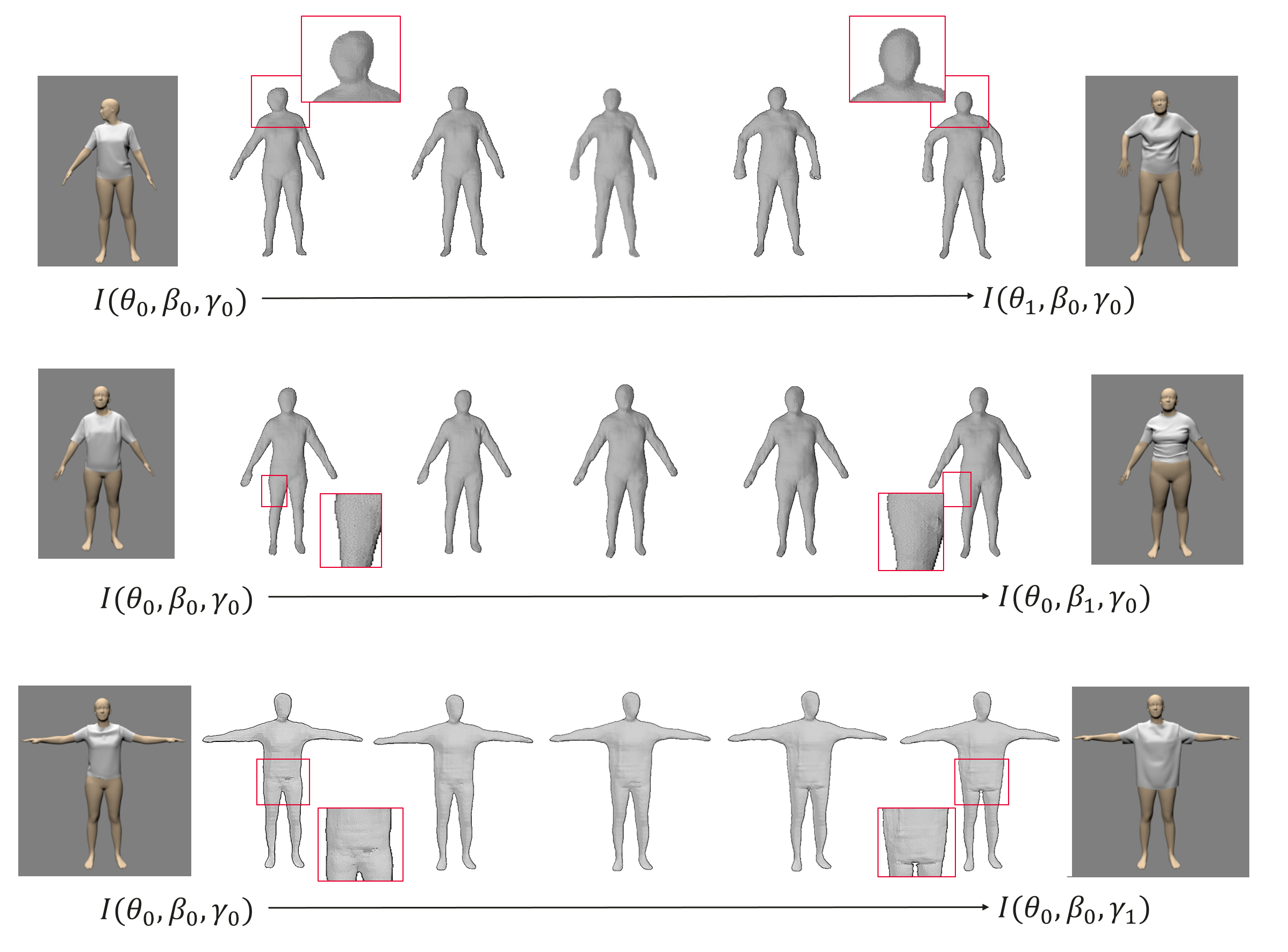}
    \caption{Example results of interpolation test: the latent code of source and target image are interpolated to reconstruct a dressed human body with new features.}
    \label{fig:interp} 
\end{figure}

Given a starting and ending frame with different features, we are able to interpolate the latent codes to achieve a smooth transition of body pose, shape or garment style in the reconstructed mesh, as shown in Fig. \ref{fig:interp}. While the pose interpolation can be used to achieve an animation effect, the body shape and garment interpolation is helpful in generating new virtual assets in an effortless way. More examples are provided in the supplementary material.

\section{Discussion and Conclusion}

In this paper, we propose the first and novel method for the task of extracting disentangled 3D attributes only from 2D image data. To illustrate the feasibility, we apply the principle to learn pose, shape, and garment representations of a human body directly from images. Our method disentangles and encodes the representations of three properties into latent codes using an encoder-decoder from the feature map of the input image and reconstructs the human body based on learned representations using the implicit function. We demonstrate the control to the reconstructed mesh by manipulating the code in the latent space. 

Although the stringent requirement on training data, where we require image pairs with only one varying property, is our limitation, we believe that our novel approach towards 2D-to-3D disentanglement can pave the way to interpretable manipulations of 3D content from 2D images alone and therefore opens the path to more seamless and controllable interaction with 3D models without the need for explicit supervision.

\newpage
\bibliographystyle{splncs04}
\bibliography{egbib}

\begin{thebibliography}{10}
\providecommand{\url}[1]{\texttt{#1}}
\providecommand{\urlprefix}{URL }
\providecommand{\doi}[1]{https://doi.org/#1}

\bibitem{alldieck2019learning}
Alldieck, T., Magnor, M., Bhatnagar, B.L., Theobalt, C., Pons-Moll, G.:
  Learning to reconstruct people in clothing from a single rgb camera. In:
  Proceedings of {IEEE} Intl. Conf. on Computer Vision and Pattern Recognition
  (CVPR) (2019)

\bibitem{anguelov2005scape}
Anguelov, D., Srinivasan, P., Koller, D., Thrun, S., Rodgers, J., Davis, J.:
  Scape: shape completion and animation of people. In: Proceedings of {ACM}
  Special Interest Group on GRAPHICS (SIGGRAPH) (2005)

\bibitem{bhatnagar2019mgn}
Bhatnagar, B.L., Tiwari, G., Theobalt, C., Pons-Moll, G.: Multi-garment net:
  Learning to dress 3d people from images. In: Proceedings of Intl. Conf. on
  Computer Vision (ICCV) (2019)

\bibitem{busam2020like}
Busam, B., Jung, H.J., Navab, N.: I like to move it: 6d pose estimation as an
  action decision process. arXiv preprint arXiv:2009.12678  (2020)

\bibitem{chabra2020deep}
Chabra, R., Lenssen, J.E., Ilg, E., Schmidt, T., Straub, J., Lovegrove, S.,
  Newcombe, R.: Deep local shapes: Learning local sdf priors for detailed 3d
  reconstruction. In: Proceedings of the European Conference on Computer Vision
  (ECCV) (2020)

\bibitem{chen2018isolating}
Chen, R.T., Li, X., Grosse, R.B., Duvenaud, D.K.: Isolating sources of
  disentanglement in variational autoencoders. In: Proceedings of Conf. on
  Neural Information Processing Systems (NeurIPS) (2018)

\bibitem{chen2016infogan}
Chen, X., Duan, Y., Houthooft, R., Schulman, J., Sutskever, I., Abbeel, P.:
  Infogan: Interpretable representation learning by information maximizing
  generative adversarial nets. In: Proceedings of Conf. on Neural Information
  Processing Systems (NeurIPS) (2016)

\bibitem{falcon2020framework}
Falcon, W., Cho, K.: A framework for contrastive self-supervised learning and
  designing a new approach. arXiv preprint arXiv:2009.00104  (2020)

\bibitem{gao2022ppp}
Gao, D., Li, Y., Ruhkamp, P., Skobleva, I., Wysocki, M., Jung, H., Wang, P.,
  Guridi, A., Busam, B.: Polarimetric pose prediction. In: European Conference
  on Computer Vision (ECCV) (October 2022)

\bibitem{gao2021information}
Gao, G., Huang, H., Fu, C., Li, Z., He, R.: Information bottleneck
  disentanglement for identity swapping. In: Proceedings of {IEEE} Intl. Conf.
  on Computer Vision and Pattern Recognition (CVPR) (2021)

\bibitem{goodfellow2014generative}
Goodfellow, I., Pouget-Abadie, J., Mirza, M., Xu, B., Warde-Farley, D., Ozair,
  S., Courville, A., Bengio, Y.: Generative adversarial nets. In: Proceedings
  of Conf. on Neural Information Processing Systems (NeurIPS) (2014)

\bibitem{gropp2020igr}
Gropp, A., Yariv, L., Haim, N., Atzmon, M., Lipman, Y.: Implicit geometric
  regularization for learning shapes. In: Proceedings of Intl. Conf. on Machine
  Learning (ICML) (2020)

\bibitem{gundogdu2019garnet}
Gundogdu, E., Constantin, V., Seifoddini, A., Dang, M., Salzmann, M., Fua, P.:
  Garnet: A two-stream network for fast and accurate 3d cloth draping. In:
  Proceedings of {IEEE} Intl. Conf. on Computer Vision and Pattern Recognition
  (CVPR) (2019)

\bibitem{hasler2009statistical}
Hasler, N., Stoll, C., Sunkel, M., Rosenhahn, B., Seidel, H.P.: A statistical
  model of human pose and body shape. In: Computer graphics forum (2009)

\bibitem{he2016deep}
He, K., Zhang, X., Ren, S., Sun, J.: Deep residual learning for image
  recognition. In: Proceedings of {IEEE} Intl. Conf. on Computer Vision and
  Pattern Recognition (CVPR) (2016)

\bibitem{higgins2016beta}
Higgins, I., Matthey, L., Pal, A., Burgess, C., Glorot, X., Botvinick, M.,
  Mohamed, S., Lerchner, A.: beta-vae: Learning basic visual concepts with a
  constrained variational framework. In: Proceedings of Intl. Conf. on Machine
  Learning (ICML) (2016)

\bibitem{Jiang2020HumanBody}
Jiang, B., Zhang, J., Cai, J., Zheng, J.: Disentangled human body embedding
  based on deep hierarchical neural network. {IEEE} Trans. on Visualization and
  Computer Graphics (TVCG)  (2020)

\bibitem{jiang2020local}
Jiang, C., Sud, A., Makadia, A., Huang, J., Nie{\ss}ner, M., Funkhouser, T.,
  et~al.: Local implicit grid representations for 3d scenes. In: Proceedings of
  {IEEE} Intl. Conf. on Computer Vision and Pattern Recognition (CVPR) (2020)

\bibitem{joo2018total}
Joo, H., Simon, T., Sheikh, Y.: Total capture: A 3d deformation model for
  tracking faces, hands, and bodies. In: Proceedings of {IEEE} Intl. Conf. on
  Computer Vision and Pattern Recognition (CVPR) (2018)

\bibitem{jung2022my}
Jung, H., Ruhkamp, P., Zhai, G., Brasch, N., Li, Y., Verdie, Y., Song, J.,
  Zhou, Y., Armagan, A., Ilic, S., et~al.: Is my depth ground-truth good
  enough? hammer--highly accurate multi-modal dataset for dense 3d scene
  regression. arXiv preprint arXiv:2205.04565  (2022)

\bibitem{kingma2013auto}
Kingma, D.P., Welling, M.: Auto-encoding variational bayes. arXiv preprint
  arXiv:1312.6114  (2013)

\bibitem{lahner2018deepwrinkles}
Lahner, Z., Cremers, D., Tung, T.: Deepwrinkles: Accurate and realistic
  clothing modeling. In: Proceedings of the European Conference on Computer
  Vision (ECCV) (2018)

\bibitem{loper2015smpl}
Loper, M., Mahmood, N., Romero, J., Pons-Moll, G., Black, M.J.: Smpl: A skinned
  multi-person linear model. {ACM} Trans. on Graphics (ToG)  (2015)

\bibitem{ma2021scale}
Ma, Q., Saito, S., Yang, J., Tang, S., Black, M.J.: {SCALE}: Modeling clothed
  humans with a surface codec of articulated local elements. In: Proceedings of
  {IEEE} Intl. Conf. on Computer Vision and Pattern Recognition (CVPR) (2021)

\bibitem{ma2020cape}
Ma, Q., Yang, J., Ranjan, A., Pujades, S., Pons-Moll, G., Tang, S., Black,
  M.J.: Learning to dress 3d people in generative clothing. In: Proceedings of
  {IEEE} Intl. Conf. on Computer Vision and Pattern Recognition (CVPR) (2020)

\bibitem{manhardt2020cps++}
Manhardt, F., Wang, G., Busam, B., Nickel, M., Meier, S., Minciullo, L., Ji,
  X., Navab, N.: Cps++: Improving class-level 6d pose and shape estimation from
  monocular images with self-supervised learning. arXiv preprint
  arXiv:2003.05848  (2020)

\bibitem{Occupancy_Networks}
Mescheder, L., Oechsle, M., Niemeyer, M., Nowozin, S., Geiger, A.: Occupancy
  networks: Learning 3d reconstruction in function space. In: Proceedings of
  {IEEE} Intl. Conf. on Computer Vision and Pattern Recognition (CVPR) (2019)

\bibitem{mu2021sdf}
Mu, J., Qiu, W., Kortylewski, A., Yuille, A., Vasconcelos, N., Wang, X.: A-sdf:
  Learning disentangled signed distance functions for articulated shape
  representation. In: Proceedings of {IEEE} Intl. Conf. on Computer Vision and
  Pattern Recognition (CVPR) (2021)

\bibitem{newell2016stacked}
Newell, A., Yang, K., Deng, J.: Stacked hourglass networks for human pose
  estimation. In: Proceedings of the European Conference on Computer Vision
  (ECCV) (2016)

\bibitem{osman2020star}
Osman, A.A.A., Bolkart, T., Black, M.J.: {STAR}: A sparse trained articulated
  human body regressor. In: Proceedings of the European Conference on Computer
  Vision (ECCV) (2020)

\bibitem{Park_2019_CVPR}
Park, J.J., Florence, P., Straub, J., Newcombe, R., Lovegrove, S.: Deepsdf:
  Learning continuous signed distance functions for shape representation. In:
  Proceedings of {IEEE} Intl. Conf. on Computer Vision and Pattern Recognition
  (CVPR) (2019)

\bibitem{patel20tailornet}
Patel, C., Liao, Z., Pons-Moll, G.: Tailornet: Predicting clothing in 3d as a
  function of human pose, shape and garment style. In: Proceedings of {IEEE}
  Intl. Conf. on Computer Vision and Pattern Recognition (CVPR) (2020)

\bibitem{Peng2020ECCV}
Peng, S., Niemeyer, M., Mescheder, L., Pollefeys, M., Geiger, A.: Convolutional
  occupancy networks. In: Proceedings of the European Conference on Computer
  Vision (ECCV) (2020)

\bibitem{pons2015dyna}
Pons-Moll, G., Romero, J., Mahmood, N., Black, M.J.: Dyna: A model of dynamic
  human shape in motion. {ACM} Trans. on Graphics (ToG)  (2015)

\bibitem{saito2019pifu}
Saito, S., Huang, Z., Natsume, R., Morishima, S., Kanazawa, A., Li, H.: Pifu:
  Pixel-aligned implicit function for high-resolution clothed human
  digitization. In: Proceedings of Intl. Conf. on Computer Vision (ICCV) (2019)

\bibitem{saito2020pifuhd}
Saito, S., Simon, T., Saragih, J., Joo, H.: Pifuhd: Multi-level pixel-aligned
  implicit function for high-resolution 3d human digitization. In: Proceedings
  of {IEEE} Intl. Conf. on Computer Vision and Pattern Recognition (CVPR)
  (2020)

\bibitem{saito2021scanimate}
Saito, S., Yang, J., Ma, Q., Black, M.J.: {SCANimate}: Weakly supervised
  learning of skinned clothed avatar networks. In: Proceedings of {IEEE} Intl.
  Conf. on Computer Vision and Pattern Recognition (CVPR) (2021)

\bibitem{sanchez2020learning}
Sanchez, E.H., Serrurier, M., Ortner, M.: Learning disentangled representations
  via mutual information estimation. In: Proceedings of the European Conference
  on Computer Vision (ECCV) (2020)

\bibitem{santesteban2021self}
Santesteban, I., Thuerey, N., Otaduy, M.A., Casas, D.: Self-supervised
  collision handling via generative 3d garment models for virtual try-on. In:
  Proceedings of {IEEE} Intl. Conf. on Computer Vision and Pattern Recognition
  (CVPR) (2021)

\bibitem{sitzmann2020implicit}
Sitzmann, V., Martel, J., Bergman, A., Lindell, D., Wetzstein, G.: Implicit
  neural representations with periodic activation functions. In: Proceedings of
  Conf. on Neural Information Processing Systems (NeurIPS) (2020)

\bibitem{wang2022phocal}
Wang, P., Jung, H., Li, Y., Shen, S., Srikanth, R.P., Garattoni, L., Meier, S.,
  Navab, N., Busam, B.: Phocal: A multi-modal dataset for category-level object
  pose estimation with photometrically challenging objects. In: Proceedings of
  the IEEE/CVF Conference on Computer Vision and Pattern Recognition. pp.
  21222--21231 (2022)

\bibitem{xu2019disn}
Xu, Q., Wang, W., Ceylan, D., Mech, R., Neumann, U.: Disn: Deep implicit
  surface network for high-quality single-view 3d reconstruction. In: Advances
  in Neural Information Processing Systems (2019)

\bibitem{yu2021pixelnerf}
Yu, A., Ye, V., Tancik, M., Kanazawa, A.: pixelnerf: Neural radiance fields
  from one or few images. In: Proceedings of {IEEE} Intl. Conf. on Computer
  Vision and Pattern Recognition (CVPR) (2021)

\bibitem{zhang2019multi}
Zhang, J., Huang, Y., Li, Y., Zhao, W., Zhang, L.: Multi-attribute transfer via
  disentangled representation. In: Proceedings of AAAI Conference on Artificial
  Intelligience(AAAI) (2019)

\bibitem{zhou20unsupervised}
Zhou, K., Bhatnagar, B.L., Pons-Moll, G.: Unsupervised shape and pose
  disentanglement for 3d meshes. In: Proceedings of the European Conference on
  Computer Vision (ECCV) (2020)

\bibitem{zhu2020deep}
Zhu, H., Cao, Y., Jin, H., Chen, W., Du, D., Wang, Z., Cui, S., Han, X.: Deep
  fashion3d: A dataset and benchmark for 3d garment reconstruction from single
  images. In: Proceedings of the European Conference on Computer Vision (ECCV)
  (2020)

\end{thebibliography}

\newpage
\appendix 
\section{Appendix}

\subsection{3D Reconstruction Loss}
We provide more details on reconstruction loss described in Section 3.3 of the main manuscript. Following \cite{saito2019pifu}, the occupancy-based 3D reconstruction loss is defined as:
 \begin{equation}
    \mathcal{L}_{recon}(\mathbf{F}, \mathbf{X}, \mathbf{M}) = \sum_{\mathbf{x} \in \mathbf{X}}\|\mathit{g}(\mathbf{F}, \mathbf{x}) - \mathbf{M}_{\scriptscriptstyle{OCC}}(\mathbf{x})\|_2^2
    \label{eq:occ}
\end{equation}
where $\mathbf{X}$ is a set of off-surface points that are sampled tightly around the surface in a way such that the half of the samples are outside the surface and the other half are inside the surface. $\mathbf{M}_{\scriptscriptstyle{OCC}}(\mathbf{x})$ is the ground truth occupancy value of $\mathbf{M}$ at 3D position $\mathbf{x}$ and $\mathbf{x} \in \mathbf{X}$.

For SDF-based reconstruction, we adopt the implicit geometric regularization \cite{gropp2020igr}, which can learn the SDF surface from point clouds without ground truth SDF supervision, as the 3D reconstruction loss. We combine the sampling strategy in \cite{sitzmann2020implicit} and \cite{saito2019pifu}: a set of points $\mathbf{X} = \mathbf{X}_{on} \cup \mathbf{X}_{off}$ are randomly sampled such that half number of points are on the surface of the mesh, denoted as $\mathbf{X}_{on}$ and the remaining points are off the surface, denoted as $\mathbf{X}_{off}$. In particular, $\mathbf{X}_{off}$ are randomly sampled around the ground-truth mesh and within the cubic space with a ratio of 3:1. For each image, the reconstruction loss $\mathcal{L}_{recon}(\mathbf{F}, \mathbf{X}, \mathbf{M})$ consists of three parts: the level set loss $\mathcal{L}_{ls}$, the Eikonal regularization loss $\mathcal{L}_{igr}$ and off surface loss $\mathcal{L}_{o}$:
\begin{align}
    \mathcal{L}_{recon}(\mathbf{F}, \mathbf{X}, \mathbf{M}) &= \lambda_{ls}\mathcal{L}_{ls} + \lambda_{igr}\mathcal{L}_{igr} +  \lambda_{o}\mathcal{L}_{o}\\
    \text{with} \quad \mathcal{L}_{ls} &= \sum_{\mathbf{x}\in \mathbf{X}_{on}}\big( \|\mathit{g}(\mathbf{F}, \mathbf{x})\|_1 + 1-\big\langle \nabla_{\mathbf{x}}\mathit{g}(\mathbf{F}, \mathbf{x}), \nabla_{\mathbf{x}}\mathbf{M}(\mathbf{x})\big\rangle \big), \\
    \mathcal{L}_{igr} &= \sum_{\mathbf{x}}\|\nabla_{\mathbf{x}}\mathit{g}(\mathbf{F}, \mathbf{x})-1\|_1,\\
    \mathcal{L}_{o} &= \sum_{\mathbf{x}\in \mathbf{X}_{off}}\exp\big(-\alpha\cdot\|g(\mathbf{F}, \mathbf{x})\|_1\big), \quad \alpha \gg 0.
    \label{eq:sdf}
\end{align}
The level set loss $\mathcal{L}_{ls}$ forces the gradient of the MLP (\ie, the predicted surface normal) $\nabla_{\mathbf{x}}\mathit{g}(\mathbf{x})$ to align with the ground-truth normal of the surface $\nabla_{\mathbf{x}}\mathbf{M}(\mathbf{x})$. The Eikonal regularization loss $\mathcal{L}_{igr}$ regularize the MLP to satisfy Eikonal equation $\|\nabla_{\mathbf{x}}\mathit{g}(\mathbf{x})\| = 1$. The off surface loss $\mathcal{L}_{o}$ push the points off the surface away from the level set surface. The overall reconstruction loss $\mathcal{L}_{recon}$ is the sum of these three components, weighted by coefficients $\lambda_{ls}$, $\lambda_{igr}$ and $\lambda_{o}$ respectively. In the experiments, the coefficients are set to $\lambda_{ls}=\lambda_{igr}=1$ and $\lambda_{o}=0.1$.

\subsection{Ablation Test on Dimensions of Latent Codes}

\label{sec:ablation}
The latent encoding plays a key role in the feature disentanglement. We observe that the dimension of the latent code is crucial for the reconstruction quality. An ablation study is carried out to justify our choice (Table \ref{tab:ablation}). We test the dimensions of latent codes of all three properties ranging from 512 to 64. Additionally, we tested the $\mathbf{l}_{\beta}\in \mathbb{R}^{16}$ as suggested by \cite{zhou20unsupervised}. The latent size is thus chosen as 128 for all properties as it leads to the lowest reconstruction error. 

\begin{table}[ht]
\begin{center}
\begin{tabular}{|llll|}
\hline
Experiment                              & Chamfer (mm) & P2S (mm) & normal (mm) \\ \hline \hline
$\mathbf{l}_{\theta},\mathbf{l}_{\beta},\mathbf{l}_{\gamma}\in \mathbb{R}^{512}$ & 4.31 & 15.75 &  10.65 \\ 
$\mathbf{l}_{\theta},\mathbf{l}_{\beta},\mathbf{l}_{\gamma}\in \mathbb{R}^{256}$  &  2.93  & 15.19  & \textbf{10.30}  \\ 
$\mathbf{l}_{\theta},\mathbf{l}_{\beta},\mathbf{l}_{\gamma}\in \mathbb{R}^{128}$ & \textbf{1.26}  & \textbf{12.12} &  10.63  \\ 
$\mathbf{l}_{\theta},\mathbf{l}_{\beta},\mathbf{l}_{\gamma}\in \mathbb{R}^{64}$ &  5.33  & 17.17 &  10.61 \\ 
$\mathbf{l}_{\beta}\in \mathbb{R}^{16}$, $\mathbf{l}_{\theta},\mathbf{l}_{\gamma}\in \mathbb{R}^{128}$ ~ & 24.05  & 38.28 &  10.55  \\ \hline
\end{tabular}
\end{center}
\caption{Accuracy of cross pose reconstruction using different latent sizes.}
\label{tab:ablation}
\end{table}

\subsection{Comparison of Direct Reconstruction between using SDF and Occupancy}
\begin{figure*}[tbh]
  \centering
  \includegraphics[width=\textwidth]{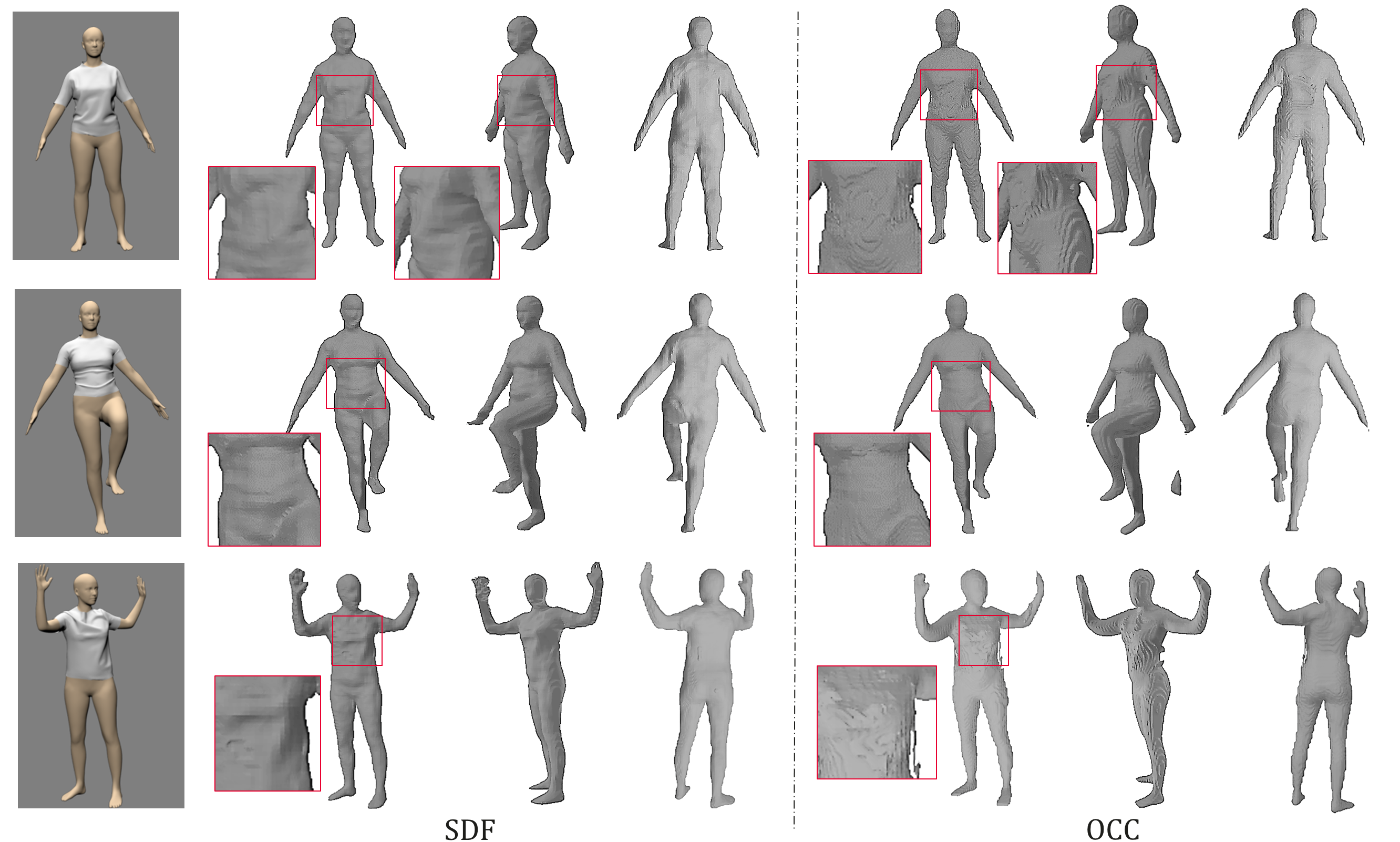}
    \caption{Qualitative comparison of direct reconstruction using SDF and occupancy.}
    \label{fig:recon} 
\end{figure*}

For the sake of reference, we additionally provide results of direct reconstruction without disentanglement. We evaluate the results of using both SDF and occupancy in the same metrics as in the main manuscript. Table \ref{tab:recon} and Fig. \ref{fig:recon} show the quantitative and qualitative comparison between the two methods. Results indicate that the direct reconstruction supervised by SDF generates a more flexible surface variation to better capture detailed wrinkles, perhaps due to the direct supervision of surface orientation. However, compared with reconstruction with disentanglement, we notice that the direct reconstruction is significantly better, especially in wrinkle details. We believe this is due to the information loss in the re-assembling of the feature map during disentanglement, which leads to degradation in high-frequency details. 

\begin{table}[tbh]

\begin{center}
\begin{tabular}{|llll|}
\hline
Experiment                              & Chamfer (mm) & P2S (mm) & normal (mm) \\ \hline\hline
Reconstruction - OCC          &  1.03   &  9.75       &  6.95     \\ 
Reconstruction - SDF                &  \textbf{0.22}  & \textbf{6.34}  & \textbf{6.69}  \\ \hline
\end{tabular}
\end{center}
\caption{Comparison of reconstruction error (median) using SDF and occupancy.}
\label{tab:recon}
\end{table}

\subsection{More Visual Examples}
\begin{figure*}
  \centering
\includegraphics[width=\textwidth]{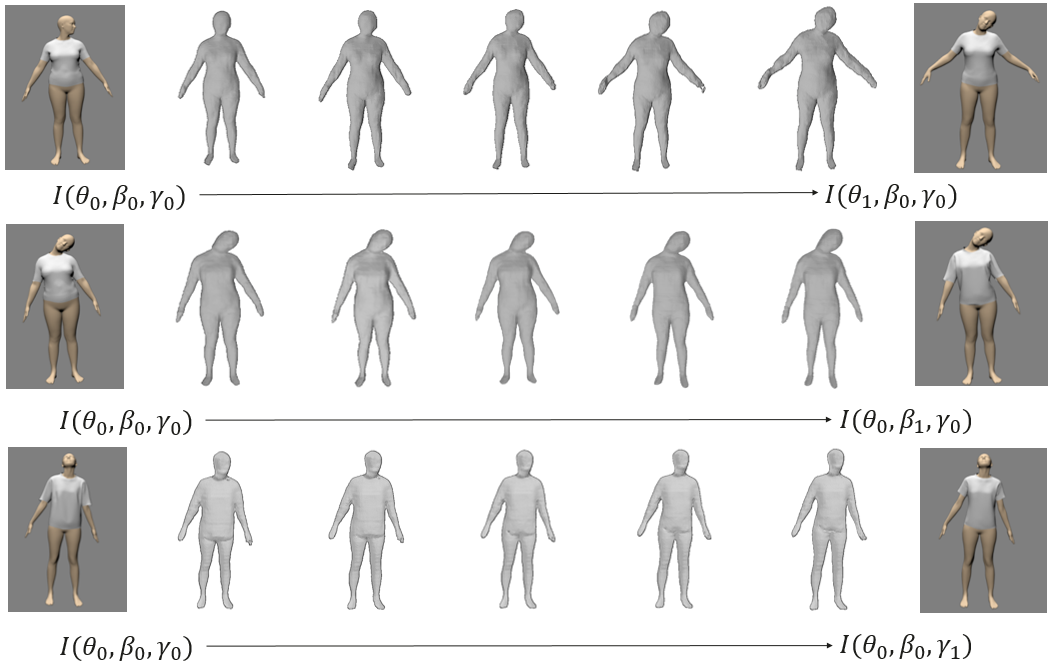}
    \caption{Random swapping of the latent codes}
    \label{fig:interp2} 
\end{figure*}

\begin{figure*}
  \centering
\includegraphics[width=\textwidth]{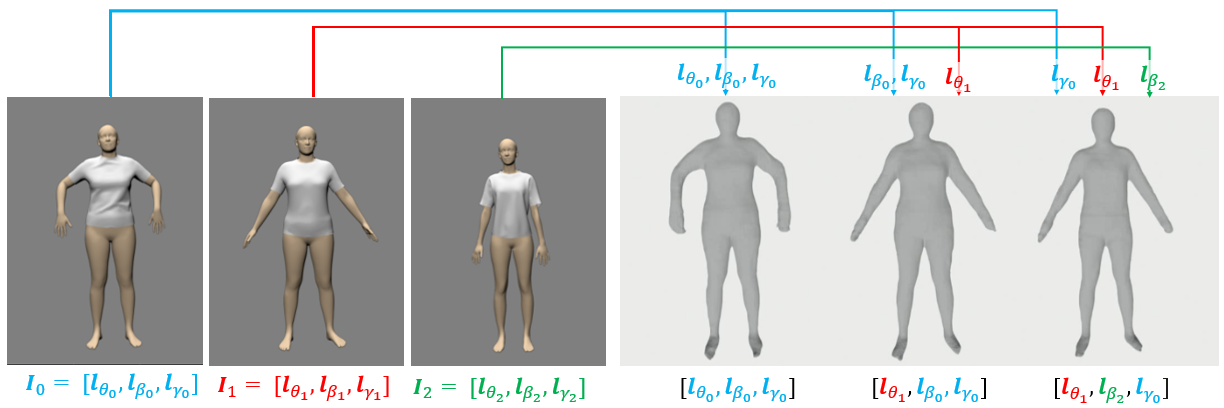}
    \caption{Random swapping of the latent codes}
    \label{fig:rand_swap2} 
\end{figure*}

We provide more visual examples in latent codes interpolation (Fig. \ref{fig:interp2}) and random swapping of latent codes (Fig. \ref{fig:rand_swap2}). 

\end{document}